\documentclass[conference]{IEEEtran}
\IEEEoverridecommandlockouts
\usepackage{cite}
\usepackage{amsmath,amssymb,amsfonts}
\usepackage{algorithmic}
\usepackage{graphicx}
\usepackage{textcomp}
\usepackage{xcolor}
\usepackage{subcaption} 
\usepackage{import}
\usepackage{color}
\usepackage[capitalise]{cleveref}
\usepackage{multirow}
\usepackage{array}
\usepackage[font=footnotesize]{caption}
\usepackage[inline]{enumitem}
\usepackage{makecell}
\usepackage{siunitx}
\usepackage{caption}
\usepackage{soul}
\usepackage{booktabs}
\usepackage[flushleft]{threeparttable}
\usepackage{float}
\usepackage{cite}
\usepackage[top=0.75in, bottom=1.05in, left=0.75in, right=0.75in]{geometry}
\usepackage{titlesec}
\titlespacing*{\subsection}{0pt}{1.0ex plus 1ex minus .2ex}{0.8ex plus .2ex}
\titlespacing*{\section}{0pt}{1.8ex plus 1ex minus .2ex}{1.0ex plus .2ex}
\def\BibTeX{{\rm B\kern-.05em{\sc i\kern-.025em b}\kern-.08em
T\kern-.1667em\lower.7ex\hbox{E}\kern-.125emX}}

\usepackage{tikz}
\newcommand\copyrighttext{%
  \footnotesize \textcopyright 2026 IEEE. Personal use of this material is permitted. Permission from IEEE must be obtained for all other uses, in any current or future media, including reprinting/republishing this material for advertising or promotional purposes, creating new collective works, for resale or redistribution to servers or lists, or reuse of any copyrighted component of this work in other works.}
\newcommand\copyrightnotice{%
\begin{tikzpicture}[remember picture,overlay]
\node[anchor=south,yshift=10pt] at (current page.south) 
  {\fbox{\parbox{\dimexpr\textwidth-\fboxsep-\fboxrule\relax}{\copyrighttext}}};
\end{tikzpicture}%
}

\begin{document}
%

\title{A GAN-Based Framework for Robust Data Synthesis in Satellite Internet Observations}

\author{\IEEEauthorblockN{
    Xiang Shi~and~Peng~Hu
\IEEEauthorblockA{
    Advanced Network and Embedded Systems Lab (AEL)\\
    Dept. of Electrical and Computer Engineering, University of Manitoba, Winnipeg, Canada}
    Email: shix2@myumanitoba.ca, peng.hu@umanitoba.ca
}
\\
\thanks{We acknowledge the support provided by the Natural Sciences and Engineering Research Council of Canada (NSERC), [funding reference number RGPIN-2022-03364] and Research Manitoba.}
}


\maketitle
\copyrightnotice
\begin{abstract}
Low-Earth orbit (LEO) satellite Internet has become an important infrastructure for enabling ubiquitous connectivity to align with the International Telecommunications Union vision for 6G telecommunications networks. However, current LEO satellite Internet observations often suffer from missing data, which complicates data augmentation task and limits the expansion of representative datasets. Given the complex characteristics of these datasets, generative AI (GenAI) presents a promising approach, yet its application in this domain has received little attention to date. In this paper, we propose a GenAI-based framework to synthesize high-fidelity data directly from incomplete LEO network observations. We propose the representative data missing scenarios, and evaluate the performance with the latest GAN- and VAE-based GenAI models on the recent WetLinks dataset. We design block-wise and point-wise missing scenarios to closely simulate the data loss that happens on real-world LEO satellite networks. Our results show the effectiveness of our proposed GAN-based framework and GT-GAN model exhibits the best performance among all models in both missing scenarios. Even under extreme conditions (e.g., 40\% of the input data is missing), GT-GAN shows the highest robustness, consistently capturing the underlying input data distribution and being the least affected in terms of generalization. Our results shed light on future directions for GenAI-based data augmentation methods and data-driven research on satellite network measurement.

\end{abstract}


%
\IEEEpeerreviewmaketitle
\vspace{-5pt}
\section{Introduction}



Low-Earth orbit (LEO) satellite Internet has become an important infrastructure that can enable ubiquitous connectivity toward the vision of 6G telecommunications networks by International Telecommunication Union (ITU). 
Recently, measurement efforts~\cite{zhao2024lens, datasetpaper, mohan2024multifaceted, casparsen2026statistical} on satellite Internet have gain increasing traction within the research community that can enable the exploration of insights and patterns of data, performance patterns, and insights of the Internet services delivered to diverse regions over the world. 

While these measurement datasets provide valuable opportunities for much-needed data-driven research on LEO satellite Internet, several fundamental challenges remain. First, existing measurements are often limited in scale and geographically sparse, hindering the generalizability of analytical results to locations without direct observations. Second, datasets collected across different regions may vary in time periods and experimental configurations, introducing inconsistencies. Third, measurements typically rely on ideal operating conditions of the measurement infrastructure (e.g., continuous Internet services without outages, satellite user terminals, dishes, and associated network devices), which are susceptible to failures or disruptions over time. These challenges can lead to missing or incomplete data samples, thereby jeopardizing the validity and robustness of subsequent data analytics or machine learning (ML) models derived from such datasets.



Our analysis of recent real-world LEO satellite measurement datasets (i.e., WetLinks\cite{datasetpaper} and LENS~\cite{zhao2024lens}) confirms the phenomenon of data missingness. Specifically, we identified 44 distinct missing-data events in the WetLinks dataset over a 5.5-month measurement period. The occurrence ranged from 33 individual gaps (i.e., a single time point of data is missing) to several consecutive data losses, with an extreme case reaching 51,570 missing points. 


When it comes to augmenting real-world measurement data, addressing missing observations is a necessary consideration. While traditional techniques like interpolation or statistical interpolation methods can successfully fill contextually consistent values to make the dataset complete. However, they focus only on filling gaps rather than generating high-fidelity synthetic data necessary for the augmentation task. ML-based generative models enhance robustness by capturing complex dependencies across both time and geographic regions. These models can identify and model hidden patterns, allowing them to reconstruct original data more accurately than traditional statistical techniques, but the reconstructed data focus only on local patterns, with no diversity data generated. Expanding on this, Generative AI (GenAI) methods, such as variational autoencoders (VAEs), generative adversarial networks (GANs), and diffusion models, can learn underlying data distributions directly from incomplete observations. Instead of merely filling gaps, these models can synthesize high-fidelity, complete synthetic data to facilitate data augmentation. They are especially effective for sparse, irregular, and high-dimensional data, making them well suited for robust synthetic data generation even when the training sources are incomplete. 

However, there is hardly any work addressing incomplete data in LEO satellite network measurements using GenAI approaches. Existing works primarily focus on general GAN-based model implementations~\cite{goodfellow2014generative, mogren2016c, smith2020conditional, yoon2019time, jeon2022gt}, while VAE-based models~\cite{kingma2013auto, desai2021timevae} and diffusion-based approaches~\cite{yuan2024diffusion} have also demonstrated strong potential in learning underlying time-series distributions. Additionally, recurrent neural network (RNN)- related models~\cite{schwarz2024interpretable} have been employed for similar tasks. However, among these GenAI frameworks, GANs are particularly well suited to LEO network data due to their ability to model complex, high-dimensional temporal dependencies. Unlike VAEs, which often produce overly smoothed results, GANs utilize a competitive training process between the generator and the discriminator. This allows GANs to capture the finer details and deeper distributions of the input time-series. Therefore, we propose a GAN-based framework to synthesize high-fidelity LEO satellite network data directly from incomplete observations, providing a robust foundation for subsequent data-driven research and network modeling. The key contributions of this paper are summarized as follows:
\begin{itemize}
    \item We propose a GAN-based framework for resolving missing data issues for the growing satellite Internet measurements. 
    \item We propose realistic missing scenarios that replicate both consecutive block-wise and stochastic point-wise data missing for commonly seen satellite Internet measurement datasets.
    \item We evaluate the framework based on the state-of-the-art GenAI models (i.e., GT-GAN, SeriesGAN, and Temporal VAE) and show that GT-GAN achieves the best performance in terms of generalization and robustness.
\end{itemize}
This paper is organized as follows: Section II reviews related work and justifies our approach and research objectives; Section III presents the problem statement and methodology; Section IV discusses the evaluation results; and Section V outlines the conclusive remarks and future work.

\section{Related Work}
Recently, GenAI approaches have shown strong promise. By learning the underlying distribution of incomplete data, these models can generate realistic and contextually consistent synthetic data, compared to traditional statistics or deterministic models. Also, GenAI-based approaches are well suited for sparse and irregular data, aligning naturally with geographically distributed measurements. By effectively capturing nonlinear and high-dimensional relationships, they offer a promising solution for handling incomplete datasets.

The introduction of the GAN framework was proposed by Goodfellow \textit{et al.}~\cite{goodfellow2014generative} in 2014. Since then, researchers have focused on continuously optimizing the framework and innovating their architectures to achieve significant progress in multiple fields. In particular, regarding the time-series tasks, a series of outstanding works has emerged. For instance, Morgan \textit{et al.}~\cite{mogren2016c} had proposed C-RNN-GAN model in 2016. This model utilized an original GAN framework to learn the distribution of time-series data and to generate new data on purpose. Both the generator and the discriminator are leveraging LSTM to address long-term memory issues in the model. Similarly, Smith \textit{et al.} proposed RCGAN~\cite{smith2020conditional} in addition to involving probability in both generator and discriminator, enabling the generation of proper data under specific conditions. In 20219, another outstanding model was proposed by Yoon \textit{et al.} named TimeGAN~\cite{yoon2019time}. This model integrates the encoder-decoder framework to ensure stable, high quality training. Furthermore, the GT-GAN model~\cite{jeon2022gt} also incorporated an autoencoder outside the GAN to improve training and generative quality. Other than that, GT-GAN mainly addressed the generation of high-fidelity synthetic data from an incomplete training dataset, without model changes. In addition to GANs, several deep learning frameworks have proven strong performance in time-series domain. For instance, Kingma \textit{et al.}~\cite{kingma2013auto} introduced the VAE, which learns the latent-space distribution of the input data to reconstruct high-fidelity fake samples. Unlike GANs, VAEs offer a more stable training process and a simpler loss function (i.e., reconstruction loss). Lastly, diffusion-based framework Diffusion-TS~\cite{yuan2024diffusion} has shown great performance in multivariate time-series generation. It leverages an encoder-decoder transformer to capture complex sequential information. Unlike traditional diffusion models that add noise to data and predict the data from the noise, Diffusion-TS directly reconstructs the data samples and uses a Fourier-based loss function to ensure both the realism and interpretability of the generated samples. 

Recent research has demonstrated the potential of GenAI in time-series data augmentation. Notably, TS-GAN~\cite{yang2023ts} has achieved significant success in augmenting the medical field by combining the LSTM and GAN frameworks to generate highly realistic synthetic time-series data. Similarly, GenAI has also proven beneficial for financial data augmentation. For instance, Dogariu \textit{et al.}~\cite{dogariu2022generation} proposed a hybrid generative model that adapts various network architectures to address the challenges of learning complex patterns and cross-correlations within financial data. Furthermore, Tang \textit{et al.}~\cite{tang2025time} augmented the energy consumption time-series data by optimizing the TimeGAN model, which subsequently improved the predictive capabilities of the hybrid CNN-GRU model.
To the best of the authors' knowledge, there are currently no studies that address how to utilize GenAI approaches to augment LEO satellite network data, particularly when the dataset is incomplete. 

\section{Problem Statement and Methodology}
To address the current research gap, we propose a GAN-based framework to examine whether existing models can capture the deep and complex distribution in LEO network measurement data and generate high-fidelity synthetic data from an incomplete dataset. To this end, our methodology is structured into four components. First, we propose two empirical missing data scenarios (i.e., block-wise and point-wise) to replicate from real-world observations. Second, we detail the generative models employed in our experiments, including GAN- and VAE-based architectures. Third, we utilize the multi-dimensional evaluation protocol proposed by Yoon \textit{et al.}~\cite{yoon2019time} to assess the quality of the synthetic data generated by each model. Finally, we dive into the experiment according to the training procedure we have designed. 

\subsection{Proposed Missing Scenario}
Based on the observed real-world missing data phenomenon. Two distinct scenarios to evaluate the models have been designed. The first scenario is block-wise missing, which replicates brief and continuous data loss caused by the Starlink dishes or satellites during system updates, resulting in consecutive gaps in the time series. The second is point-wise missing, representing random data loss at specific time points where all features become unavailable. This design is also informed by the real-world investigations of the representative satellite network measurement datasets. In the example of WetLinks dataset, we have identified 44 instances of missing data. These range from 33 single-point gaps (i.e., point-wise) to multiple consecutive gaps (i.e., block-wise), including six gaps for two consecutive missing data points and several gaps spanning more than two data points, with extreme cases of up to 51,570 consecutive gaps. By replicating these observed real-world missing patterns, we can effectively simulate and evaluate the quality of the model's synthetic data when partial input data is missing.

Block-wise missing scenario is used to assess the quality of synthetic data generated by models when continuous data is lost in the satellite network dataset. The original dataset is divided into several sliding windows of fixed length $L$. For the sequence of data points $X = {x_1, x_2, .... , x_L}$, we define a mask vector $M = {m_1, m_2, .... , m_L}$, where $m_i \in \{0, 1\}$ indicates the missingness of the data. The first $k$ observations of each window are masked, where $k \in \{5, 10, 20, 40\}$ represent different scales of missing observations. The formal definition is as follows:
\vspace{-5pt}
\begin{equation}
    m_i = 
    \begin{cases} 
    0, & (1 \leq t \leq k) \\
    1, & (k < t \leq L)
    \end{cases}
    \text{ , where }k \in \{5, 10, 20, 40\}
\end{equation}




Point-wise missing scenario is used to simulate sensor or system outages where all features are lost at specific time points. In this design, the missingness depends solely on the time index $t$. For each timestep, a mask value $m_t$ is drawn from a Bernoulli distribution, where the mask for each element is defined as $M(t,i)= m_t$ for all features $i \in \{1, \dots,D\}$. We defined four missing data proportions, with $p \in \{5\%, 10\%, 20\%, 40\%\}$ representing different missing rates, while maintaining the fixed random seed for reproducibility. The formal definition is as follows:

\vspace{-4pt}
\begin{equation}
\begin{gathered}
    m_{t} \sim Bernoulli(1 - p)\\
    \text{where } p \in \{5\%, 10\%, 20\%, 40\%\}
\end{gathered}
\end{equation}


\subsection{Model Selection} 
We utilize three different generative models to compare their performance in synthesizing LEO satellite network data. GT-GAN~\cite{jeon2022gt} and SeriesGAN~\cite{eskandarinasab2024seriesgan} were selected as the primary GAN-based models due to their demonstrated ability to capture the complex, non-linear temporal dependencies in the underlying data distribution. To provide a robust baseline, we implement a Temporal VAE based on the VAE framework originally proposed by Kingma \textit{et al.} \cite{kingma2013auto}. The Temporal VAE architecture consists of a 2-layer bidirectional Gated Recurrent Units (GRU) encoder and a 2-layer GRU decoder, both with a hidden dimension size of 64. The latent space dimension size is set to 32. To handle data sparsity, we integrate a binary mask $M$ into the loss function during training. This ensures that the reconstruction loss is backpropagated only through observed data points, preventing the model from being biased by missing samples. 

\subsection{Multi-dimensional Evaluation Framework}
To assess the quality of the generated data, our evaluation framework involves quantitative and qualitative metrics. For the quantitative metric, we leverage the discriminative and predictive scores as proposed in TimeGAN~\cite{yoon2019time}. The discriminative score measures the similarity between the distribution of real and synthetic data. Train a 2-layer RNN classifier model using the real data, and use the model to recognize whether the data is synthetic. The lower discriminative score indicates the model has successfully captured the underlying distribution, generating the synthetic data indistinguishable from the real data. For the predictive score, we adopt the Train-on-Synthetic, Test-on-Real (TSTR) evaluation protocol. By using the synthetic data to train a post-hoc sequence prediction model (i.e., a 2-layer RNN) to predict the future samples, and then evaluating the prediction model on the real dataset. The score is calculated as $|1 - MAE|$, where the value closer to 0 indicates better performance. The metric objective is to compare with the real dataset whether the synthetic data can still support accurate predictions of future events. For the qualitative metric, we employ t-SNE~\cite{van2008visualizing} to project the high-dimensional temporal sequence into a two-dimensional space, enabling visual comparison of the overall structures and local patterns between the real and synthetic data distributions. 

\subsection{Training Procedure}
To ensure a fair assessment of each model's performance across different missing data scenarios, our experimental pipeline follows three structured steps. First, based on the predefined missing data scenarios, we transform an original dataset (e.g, WetLinks) into multiple incomplete versions for training. Second, each model is fine-tuned on these specific incomplete datasets to ensure stable performance. Finally, we utilize the multi-dimensional evaluation protocol to assess all three models' performance and quality of the generated synthetic data. 

\section{Evaluation}
In this section, we conduct a comparative study across two GAN-based models (i.e., GT-GAN and SeriesGAN) and one VAE-based model (i.e., Temporal VAE). Our objective is to comprehensively examine the generalization and robustness of these three models in incomplete LEO satellite network data.


\subsection{Experimental Dataset}
We utilized a 2-day subset (i.e., 2,880 samples) from the Starlink dataset for the WetLinks Enschede region~\cite {datasetpaper}, spanning October 7-8, 2023. This high-dimensional dataset includes eight network and spatial features (e.g., throughput, latency-related, and azimuth/elevation). Two minor point-wise gaps were handled manually via mean imputation. Unlike purely performance-focused datasets, WetLinks provides a combination of performance metrics and spatial coordinates, enabling our evaluation framework to assess whether GenAI models can capture the deep dependencies between satellite movement and network quality.


\subsection{Comparison of Block-wise Missing Scenario}
In the scenario of block-wise missing data, we conducted comparative experiments on three models across varying scales of data loss (i.e., $k$). The results of the quantitative metric are shown in Table~\ref{tab:blockwise_table_results}. In terms of overall scores, GT-GAN is the top-performing model of the three. GT-GAN model's discriminative score ranges from approximately 0.4 to 0.48, which is significantly lower than that of SeriesGAN and Temporal VAE. This indicates that the synthetic data generated by GT-GAN is more difficult for the RNN classifier to distinguish from real data (i.e., it is more similar to real data). For the predictive scores, GT-GAN is also the best-performing model among the three. In all four $k$ experiments, the GT-GAN model's predictive scores were lower than those of the other two models. This suggests that the GT-GAN model generated synthetic data not only statistically similar to real data but also useful for downstream data analysis tasks.

\begin{figure}[htbp]
     \centering
     \begin{subfigure}[b]{0.23\textwidth}
         \centering
         \includegraphics[width=\textwidth]{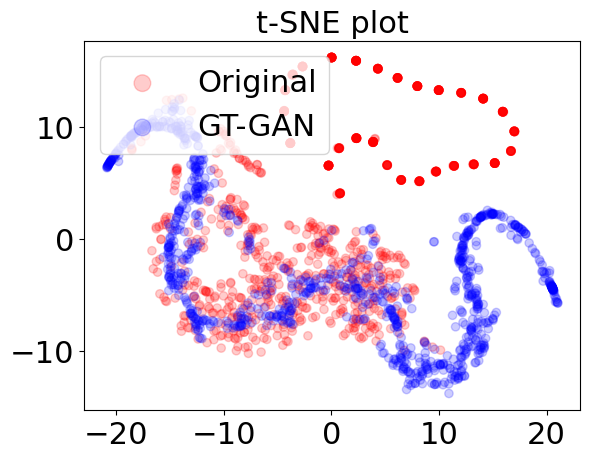}
         \caption{$k = 5$}
     \end{subfigure}
     \hfill
     \begin{subfigure}[b]{0.23\textwidth}
         \centering
         \includegraphics[width=\textwidth]{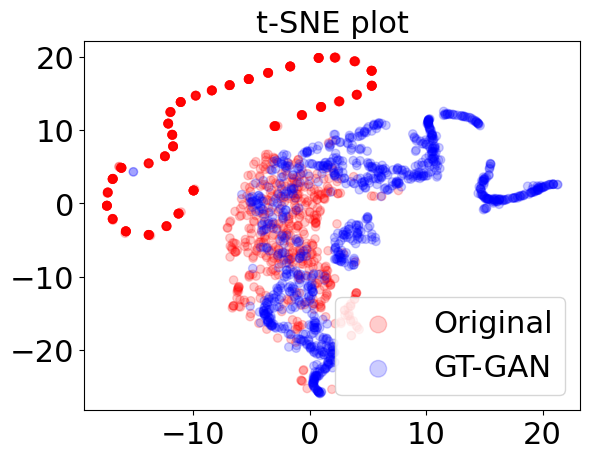}
         \caption{$k = 10$}
     \end{subfigure}
     
     \begin{subfigure}[b]{0.23\textwidth}
         \centering
         \includegraphics[width=\textwidth]{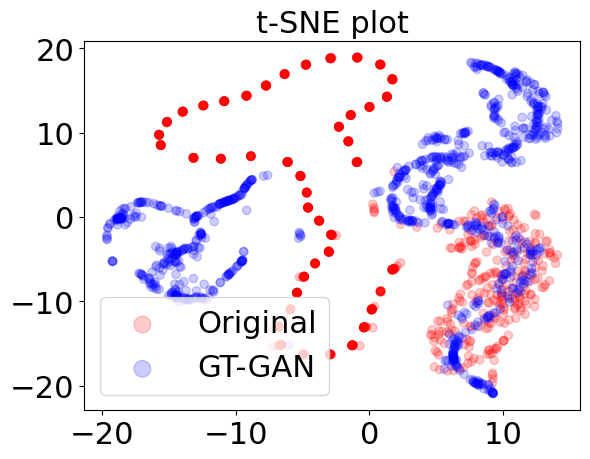}
         \caption{$k = 20$}
     \end{subfigure}
     \hfill
     \begin{subfigure}[b]{0.23\textwidth}
         \centering
         \includegraphics[width=\textwidth]{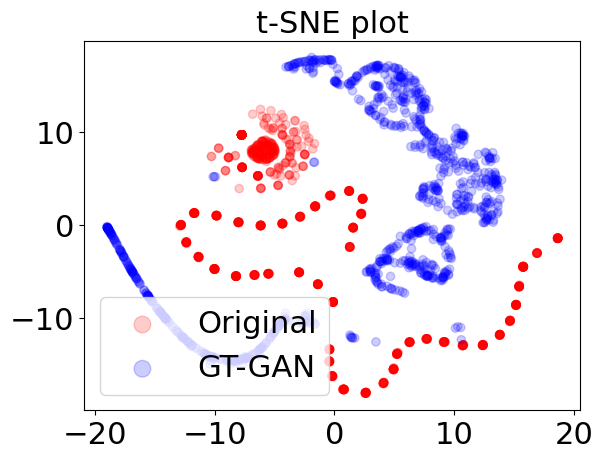}
         \caption{$k = 40$}
     
     \end{subfigure}
     \caption{GT-GAN multiple t-SNE plots for block-wise missing comparison}
     \label{fig:blockwise_gtgan_tsne}
\end{figure}

\begin{figure}[htbp]
     \centering
     \begin{subfigure}[b]{0.23\textwidth}
         \centering
         \includegraphics[width=\textwidth]{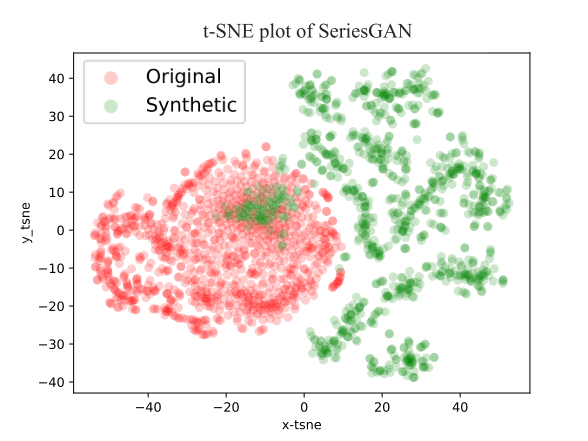}
         \caption{$k = 5$}
     \end{subfigure}
     \hfill
     \begin{subfigure}[b]{0.23\textwidth}
         \centering
         \includegraphics[width=\textwidth]{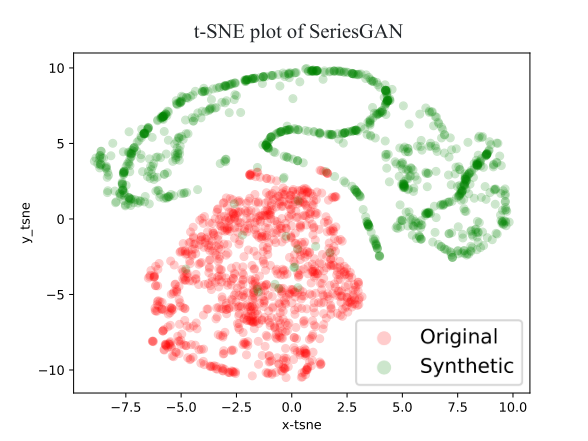}
         \caption{$k = 10$}
     \end{subfigure}
     
     \begin{subfigure}[b]{0.23\textwidth}
         \centering
         \includegraphics[width=\textwidth]{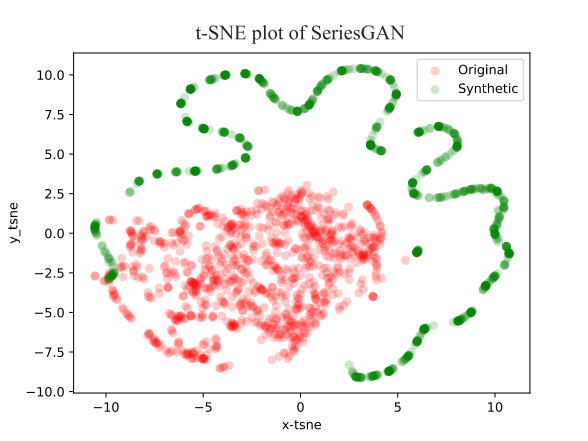}
         \caption{$k = 20$}
     \end{subfigure}
     \hfill
     \begin{subfigure}[b]{0.23\textwidth}
         \centering
         \includegraphics[width=\textwidth]{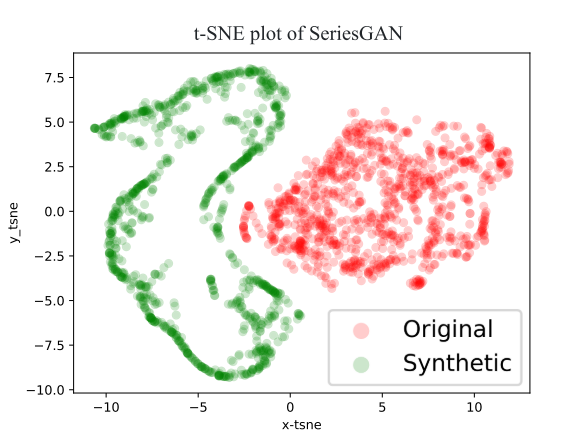}
         \caption{$k = 40$}
     
     \end{subfigure}
     \caption{SeriesGAN multiple t-SNE plots for block-wise missing comparison}
     \label{fig:blockwise_seriesGAN_tsne}
\end{figure}

\begin{figure}[htbp]
     \centering
     \begin{subfigure}[b]{0.23\textwidth}
         \centering
         \includegraphics[width=\textwidth]{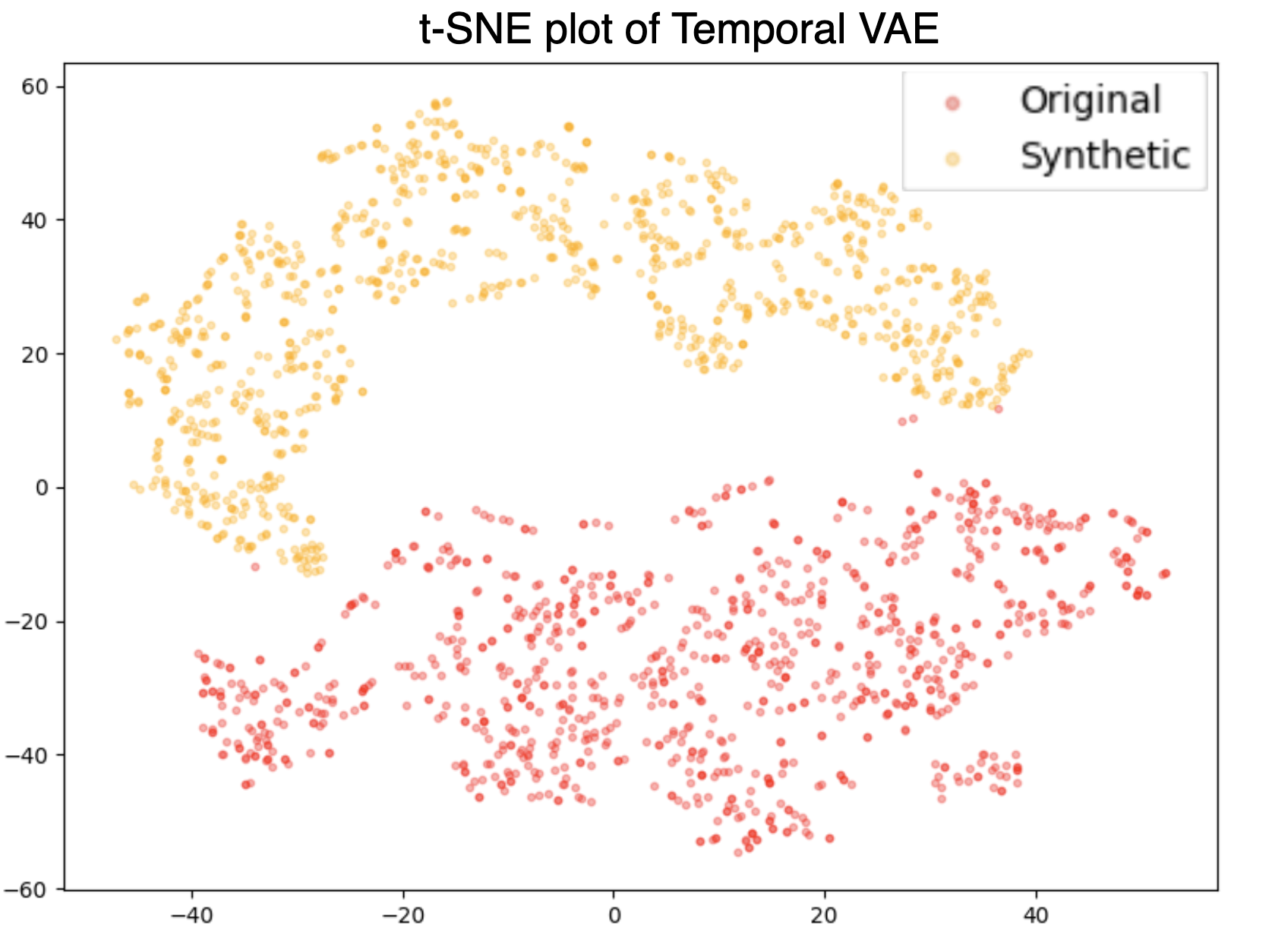}
         \caption{$k = 5$}
     \end{subfigure}
     \hfill
     \begin{subfigure}[b]{0.23\textwidth}
         \centering
         \includegraphics[width=\textwidth]{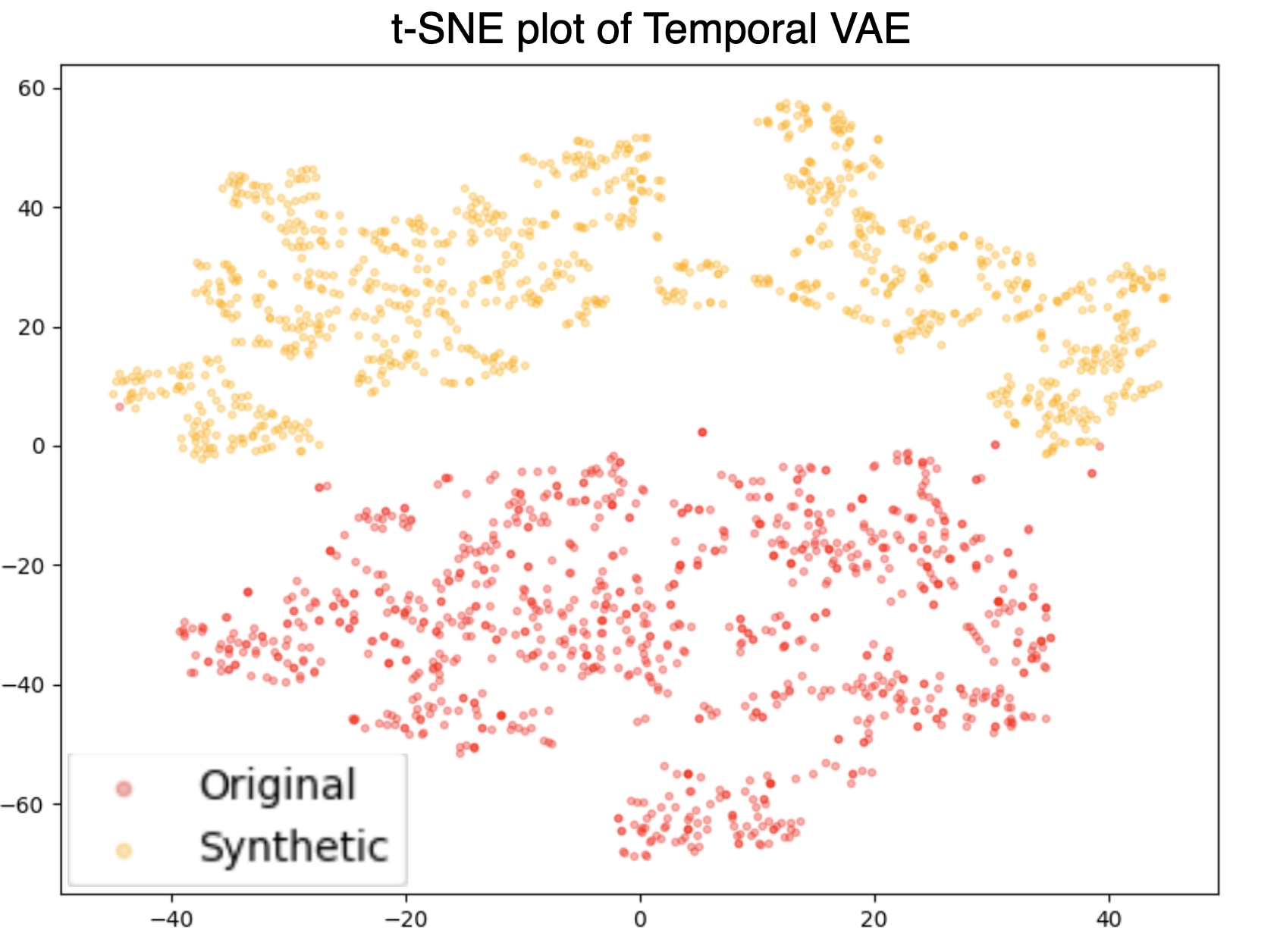}
         \caption{$k = 10$}
     \end{subfigure}
     
     \begin{subfigure}[b]{0.23\textwidth}
         \centering
         \includegraphics[width=\textwidth]{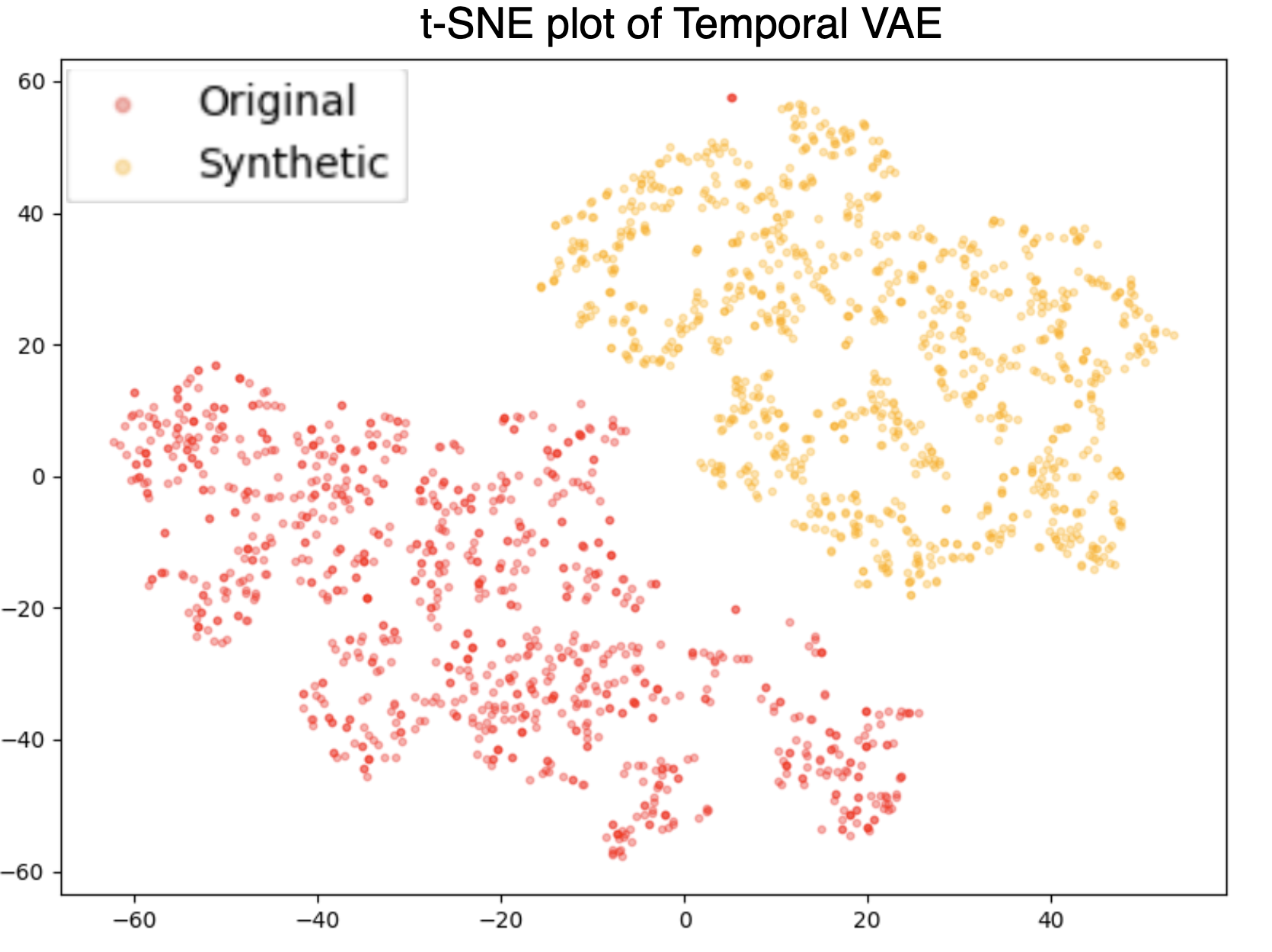}
         \caption{$k = 20$}
     \end{subfigure}
     \hfill
     \begin{subfigure}[b]{0.23\textwidth}
         \centering
         \includegraphics[width=\textwidth]{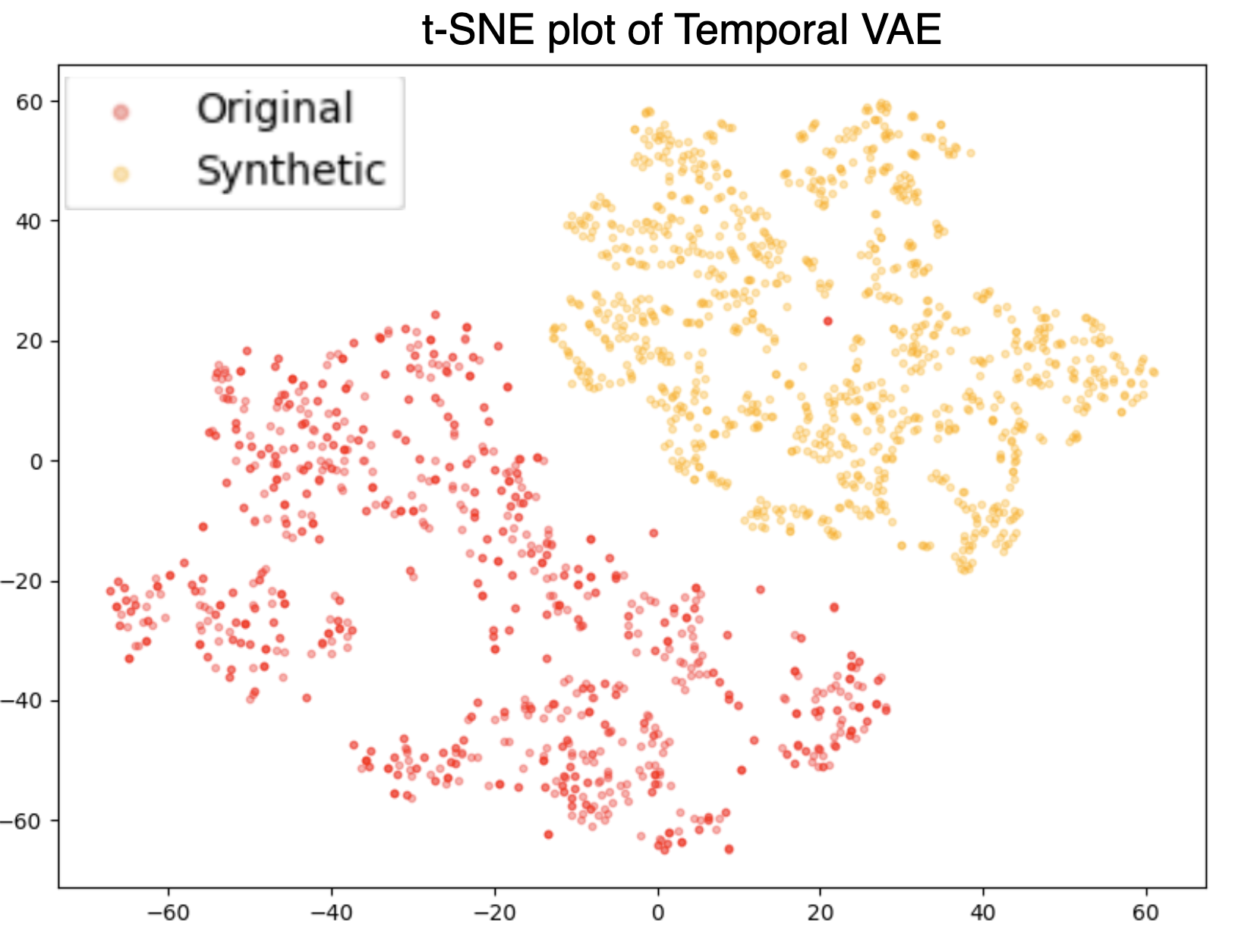}
         \caption{$k = 40$}
     \end{subfigure}
     \caption{Temporal VAE model multiple t-SNE plots for block-wise missing scenario comparison}
     \label{fig:blockwise_temporalVAE_tsne}
\end{figure}

\begin{table}[htbp]
    \centering
    \caption{Comparison of Generative Models on WetLinks dataset under Block-wise Missing Scenarios}
    \label{tab:blockwise_table_results}
    \begin{tabular}{c|c|c|c|c}
         \toprule
         $k$    & \textbf{Metric} & \textbf{GT-GAN} & \textbf{SeriesGAN} & \textbf{Temporal VAE} \\
         \hline
         \multirow{2}{*}{5} & Discriminative & \textbf{0.4054} & 0.4904 &  0.4866\\
                            & Predictive & \textbf{0.0544} & 0.0675 & 0.0875\\
         
         \multirow{2}{*}{10} & Discriminative & 0.4339 &  0.4989 &  \textbf{0.3184}\\
                             & Predictive & \textbf{0.0457} & 0.0673 & 0.1184\\

         \multirow{2}{*}{20} & Discriminative & 0.4825 &  0.4927 &  \textbf{0.4122}\\
                             & Predictive & \textbf{0.0509} & 0.0698 & 0.1845\\

         \multirow{2}{*}{40} & Discriminative & \textbf{0.4005} &  0.5 &  0.5\\
                             & Predictive & \textbf{0.0458} & 0.0768 & 0.3223\\
         \bottomrule
    \end{tabular}
    \begin{tablenotes}
        \item \textit{Note:} Lower scores in both metrics indicate better performance. Bold values represent the best results for each missing case ($k$).
    \end{tablenotes}
\end{table}
We provide each model's t-SNE plots in Fig.~\ref{fig:blockwise_gtgan_tsne}, Fig.~\ref{fig:blockwise_seriesGAN_tsne}, and Fig.~\ref{fig:blockwise_temporalVAE_tsne}. The core principle of t-SNE is that the higher the overlap between the generated data points and the real data points, the stronger the model's ability to capture the underlying data distribution. From the GT-GAN model's t-SNE plots (Fig.~\ref{fig:blockwise_gtgan_tsne}), we found that when data is missing at $k=5$ and $k=10$, the blue (synthetic) and red (real) data points overlap clearly. As $k$ increases to 40, although the blue points begin to deviate slightly, they still attempt to cover the cluster of the red points. This confirms the conclusion that GT-GAN mostly has the lowest discriminative and predictive scores in the quantitative metrics. For the t-SNE plots of the SeriesGAN (Fig.~\ref{fig:blockwise_seriesGAN_tsne}) and Temporal VAE models (Fig.~\ref{fig:blockwise_temporalVAE_tsne}), we found that in most cases of $k$, both models' generated data points did not overlap with real data points. In summary, GT-GAN consistently outperforms the other two models, producing stronger performance across both quantitative and qualitative metrics. 

\subsection{Comparison of Point-wise Missing Scenario}
{In the point-wise missing data scenario, the quantitative metrics of the three models are shown in Table~\ref{tab:pointwise_table_result}. When $p\le10\%$ (i.e., low missing rate), the input sequences still contain the short-term patterns and their local temporal continuity. Under this condition, Temporal VAE and SeriesGAN achieve the best discriminative and predictive scores, respectively. This is because Temporal VAE relies on a reconstruction-based training objective, which is effective at generating indistinguishable synthetic data that achieves high discriminative scores. Similarly, SeriesGAN can better model short-term patterns under slightly missing sequences. In contrast, GT-GAN is designed to learn more global sequence-level features and to remain robust when the data becomes more fragmented. Therefore, when the missing rate is low, its advantage is not fully reflected, and it isn't fit local point-wise details as well as Temporal VAE or SeriesGAN. However, as the missing rate increases to $p \ge 20\%$, the point-wise missing values make the sequences more fragmented and weaken the local temporal continuity. In this case, SeriesGAN and Temporal VAE become more affected by the lack of local continuous context, leading to degraded performance. GT-GAN remains more stable because it can extract global temporal features from incomplete sequences. Even when 40\% of the input data is missing, GT-GAN still maintains stable generalization, demonstrating greater robustness under severe point-wise missingness.}


\begin{table}[htbp]
    \centering
    \caption{Comparison of Generative Models on WetLinks dataset under Point-wise Missing Scenarios}
    \label{tab:pointwise_table_result}
    \begin{tabular}{c|c|c|c|c}
         \toprule
         \textbf{$p$} & \textbf{Metric} & \textbf{GT-GAN} & \textbf{SeriesGAN} & \textbf{Temporal VAE} \\
         \hline
         \multirow{2}{*}{5\%} & Discriminative & 0.4726 & 0.4966 &  \textbf{0.4107}\\
                            & Predictive & 0.0592 & \textbf{0.0486} & 0.0697\\
         
         \multirow{2}{*}{10\%} & Discriminative & 0.4904 &  0.4986 &  \textbf{0.3541}\\
                             & Predictive & 0.0664 & \textbf{0.0488} & 0.1062\\

         \multirow{2}{*}{20\%} & Discriminative & \textbf{0.4409} &  0.5 &  0.4910\\
                             & Predictive & \textbf{0.0463} & 0.2502 & 0.1462\\

         \multirow{2}{*}{40\%} & Discriminative & \textbf{0.4328} &  0.5 &  0.5\\
                             & Predictive & \textbf{0.0502} & 0.3647 & 0.2194\\
         \bottomrule
    \end{tabular}
    \begin{tablenotes}
        \item \textit{Note:} Lower scores in both metrics indicate better performance. Bold values represent the best results for each missing rate ($p$).
    \end{tablenotes}
\end{table}

\begin{figure}[htbp]
     \centering
     \begin{subfigure}[b]{0.23\textwidth}
         \centering
         \includegraphics[width=\textwidth]{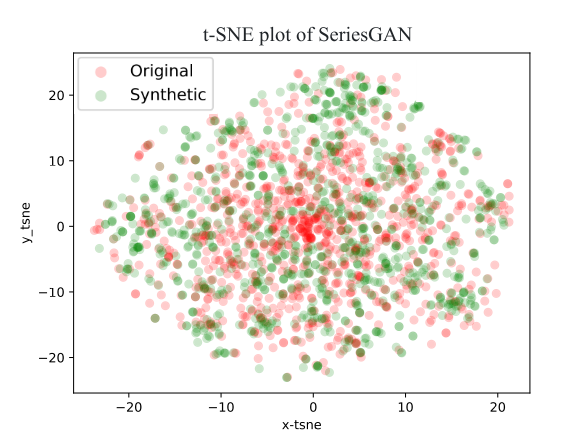}
         \caption{$p = 5\%$}
     \end{subfigure}
     \hfill
     \begin{subfigure}[b]{0.23\textwidth}
         \centering
         \includegraphics[width=\textwidth]{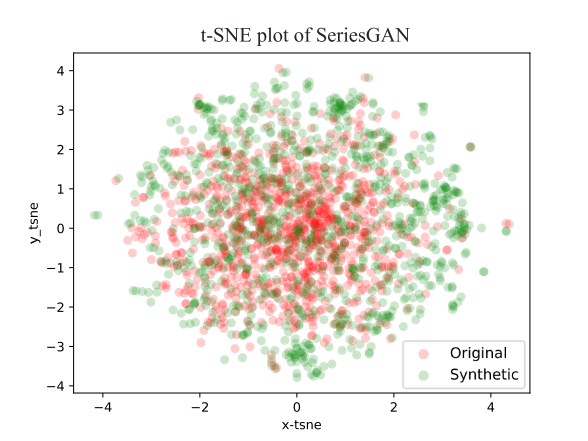}
         \caption{$p = 10\%$}
     \end{subfigure}
     
     \begin{subfigure}[b]{0.23\textwidth}
         \centering
         \includegraphics[width=\textwidth]{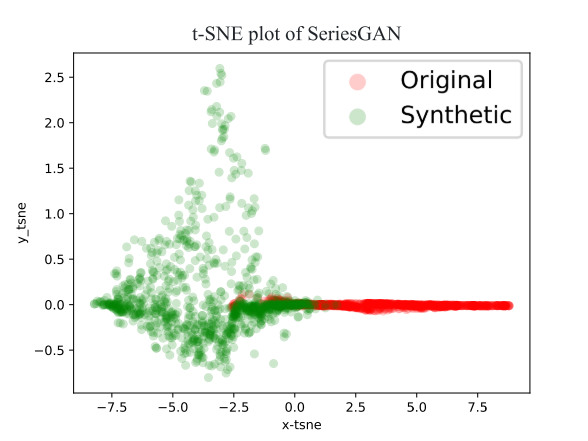}
         \caption{$p=20\%$}
     \end{subfigure}
     \hfill
     \begin{subfigure}[b]{0.23\textwidth}
         \centering
         \includegraphics[width=\textwidth]{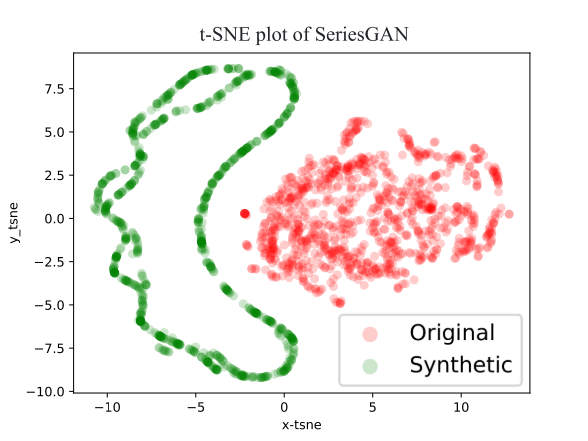}
         \caption{$p=40\%$}
     \end{subfigure}
     \caption{SeriesGAN multiple t-SNE plots for Point-wise missing scenario comparison}
     \label{fig:S2_SeriesGAN_plots}
\end{figure}

\begin{figure}[htbp]
     \centering
     \begin{subfigure}[b]{0.23\textwidth}
         \centering
         \includegraphics[width=\textwidth]{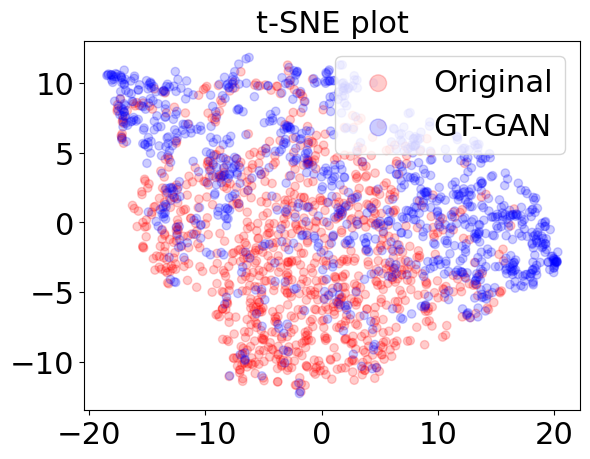}
         \caption{$p = 5\%$}
     \end{subfigure}
     \hfill
     \begin{subfigure}[b]{0.23\textwidth}
         \centering
         \includegraphics[width=\textwidth]{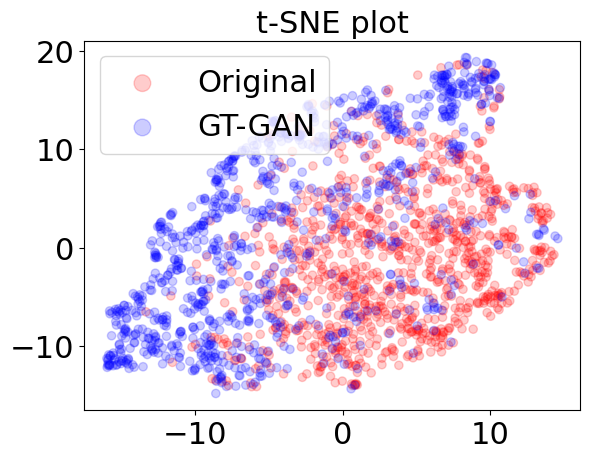}
         \caption{$p = 10\%$}
     \end{subfigure}
     
     \begin{subfigure}[b]{0.23\textwidth}
         \centering
         \includegraphics[width=\textwidth]{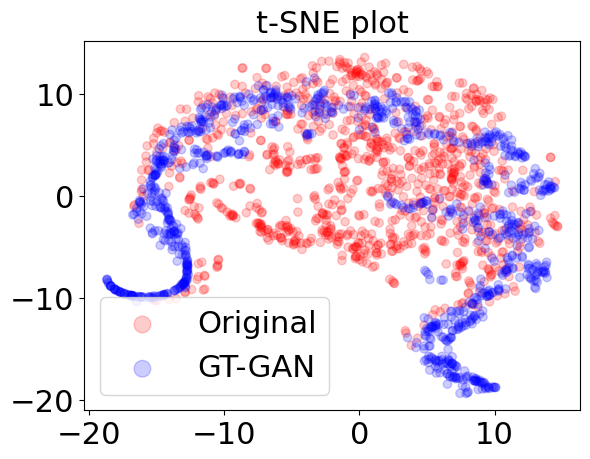}
         \caption{$p=20\%$}
     \end{subfigure}
     \hfill
     \begin{subfigure}[b]{0.23\textwidth}
         \centering
         \includegraphics[width=\textwidth]{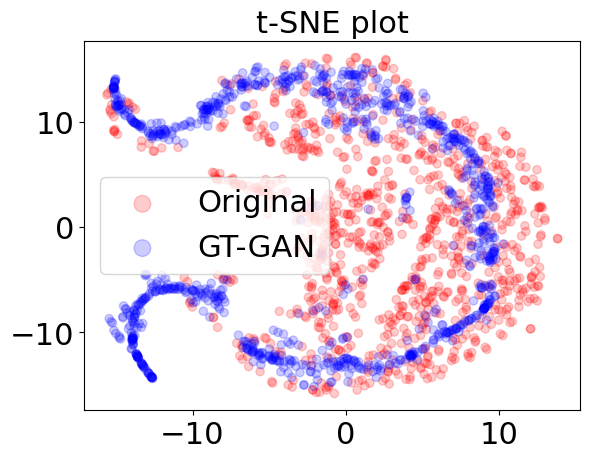}
         \caption{$p=40\%$}
     \end{subfigure}
     \caption{GT-GAN model multiple t-SNE plots for Point-wise missing scenario comparison}
     \label{fig:S2_GT-GAN_plots}
\end{figure}

\begin{figure}[htbp]
     \centering
     \begin{subfigure}[b]{0.23\textwidth}
         \centering
         \includegraphics[width=\textwidth]{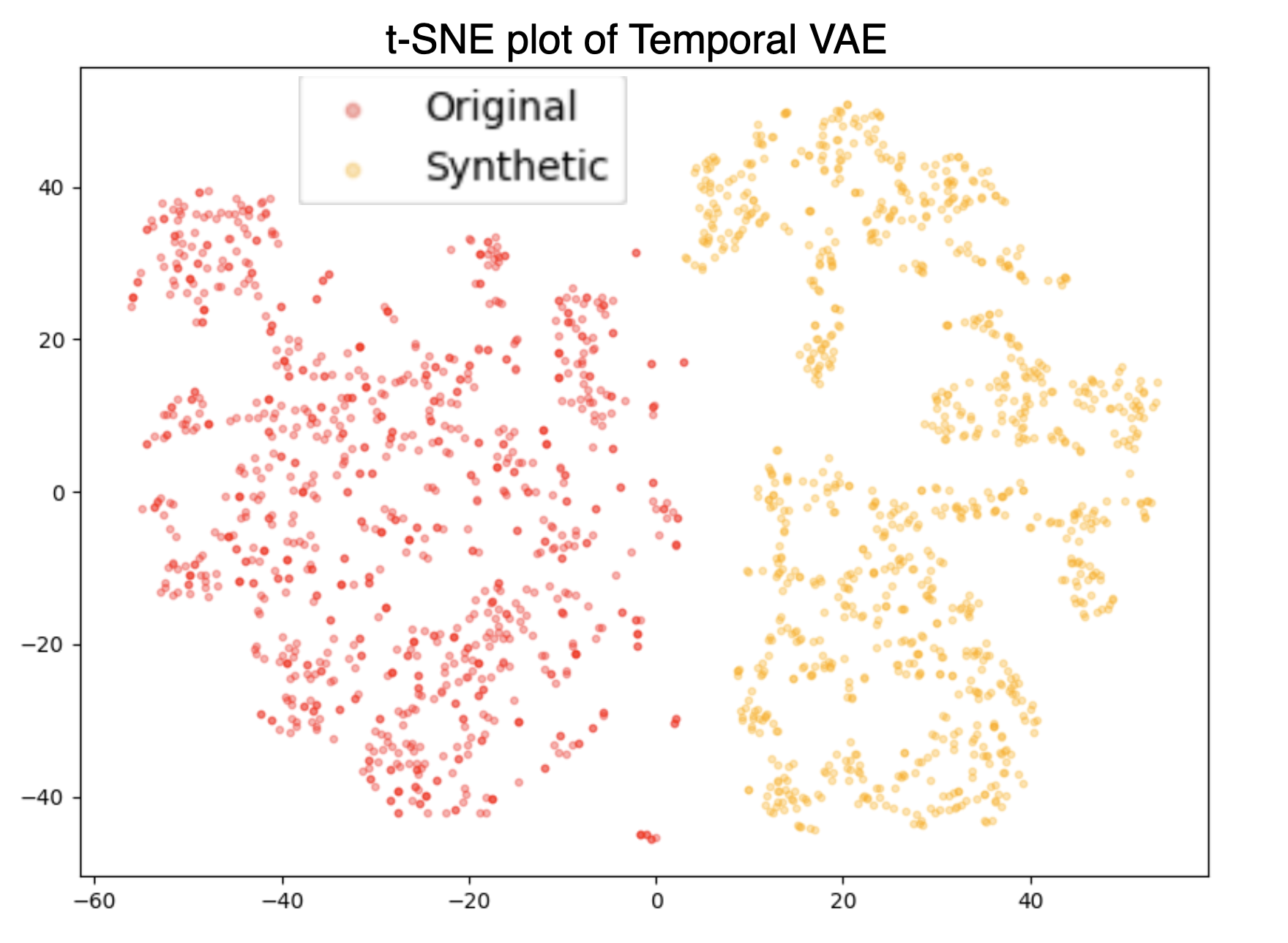}
         \caption{$p = 5\%$}
     \end{subfigure}
     \hfill
     \begin{subfigure}[b]{0.23\textwidth}
         \centering
         \includegraphics[width=\textwidth]{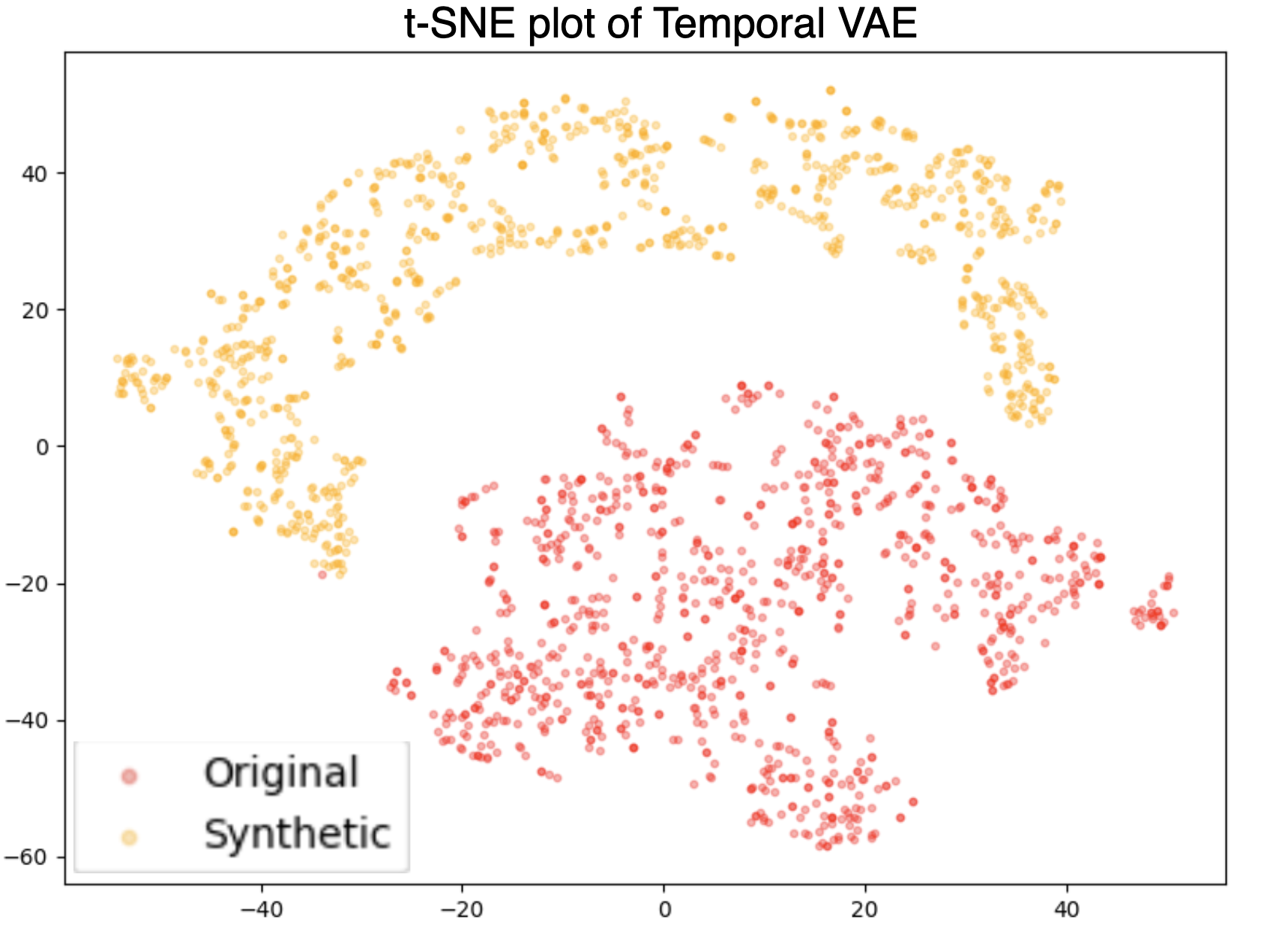}
         \caption{$p = 10\%$}
     \end{subfigure}
     
     \begin{subfigure}[b]{0.23\textwidth}
         \centering
         \includegraphics[width=\textwidth]{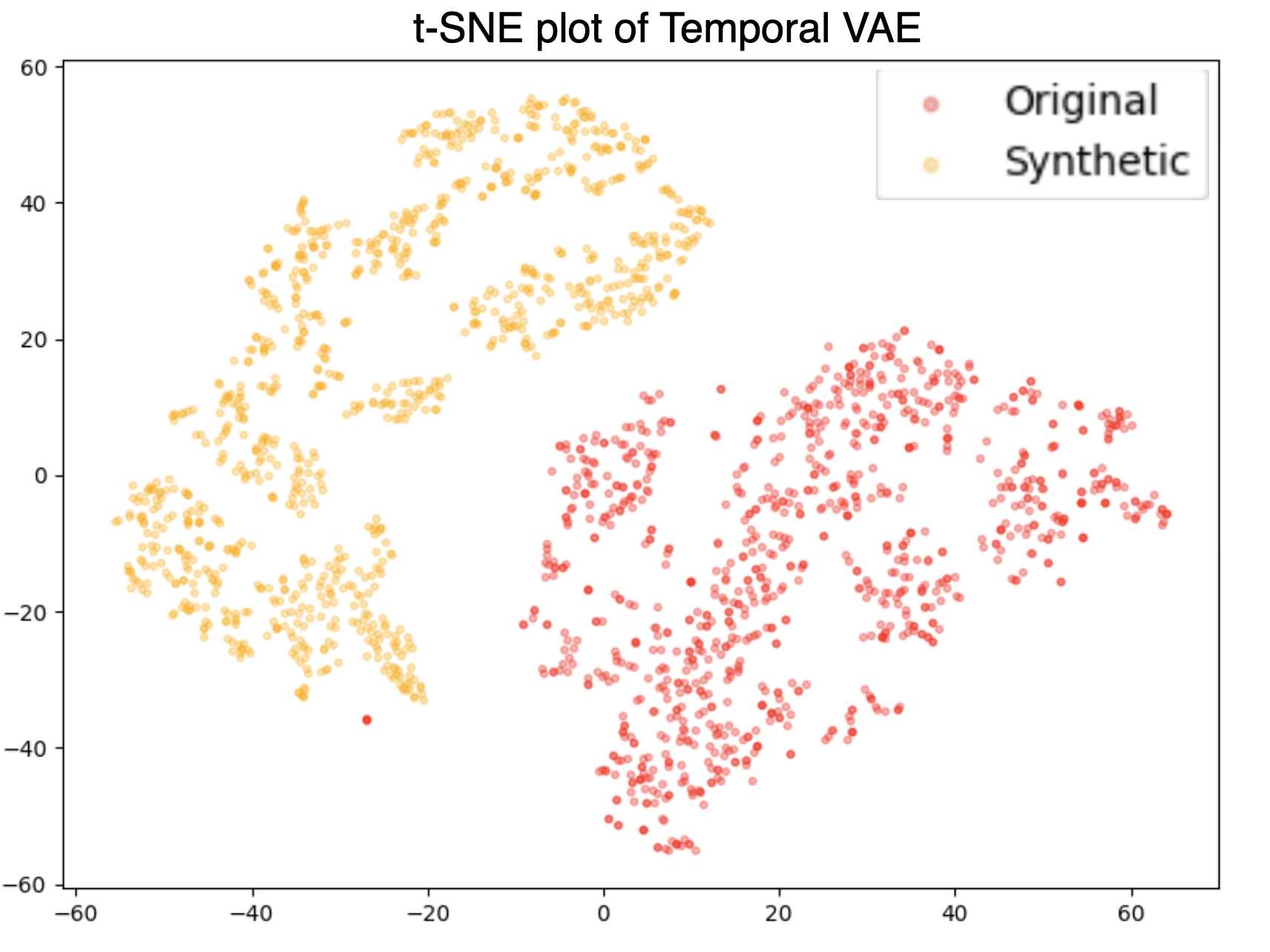}
         \caption{$p=20\%$}
     \end{subfigure}
     \hfill
     \begin{subfigure}[b]{0.23\textwidth}
         \centering
         \includegraphics[width=\textwidth]{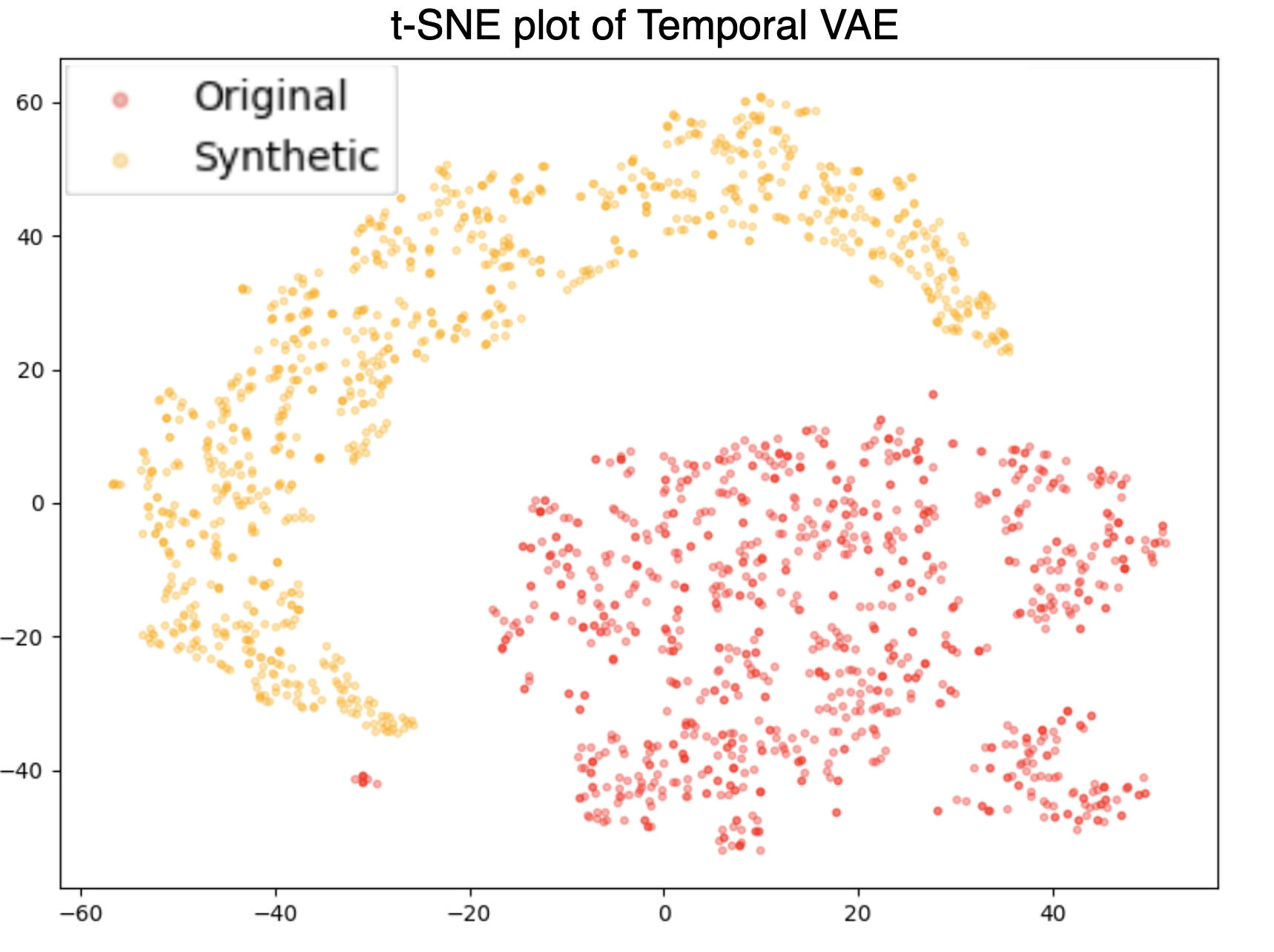}
         \caption{$p=40\%$}
     \end{subfigure}
     \caption{Temporal VAE model multiple t-SNE plots for point-wise missing scenario comparison}
     \label{fig:S2_TemporalVAE_plots}
\end{figure}

{The t-SNE visualizations effectively validate the quantitative findings. At lower missing rates (i.e., $p\le10\%$), both SeriesGAN (Figs.~\ref{fig:S2_SeriesGAN_plots}a and~\ref{fig:S2_SeriesGAN_plots}b) and GT-GAN  (Figs.~\ref{fig:S2_GT-GAN_plots}a and~\ref{fig:S2_GT-GAN_plots}b) demonstrate strong performance, generating synthetic data that visually aligns well with the underlying real data patterns. Temporal VAE (Fig.~\ref{fig:S2_TemporalVAE_plots}) also maintains similar structure density despite clearer cluster separation. However, a distinct performance gap emerges as the missing rate increases to $p\ge20\%$. As evidenced by the t-SNE plots (Figs.~\ref{fig:S2_SeriesGAN_plots}c and~\ref{fig:S2_SeriesGAN_plots}d), SeriesGAN suffers from severe mode collapse. Instead of capturing the full diversity of the training data distribution, the model's outputs converged to narrow ranges of repetitive patterns, resulting in a complete lack of overlap between the synthetic and real data points. This failure occurs because, in GAN architectures, highly incomplete inputs often cause the discriminator to overpower the generator, leading to gradient vanishing. Conversely, although we also observed that mode collapse happened in the GT-GAN model's t-SNE plots when $p\ge20\%$ (i.e., Figs.~\ref{fig:S2_GT-GAN_plots}c and~\ref{fig:S2_GT-GAN_plots}d). However, its synthetic points still follow the underlying distribution and partially overlap with the real data. This sharp contrast further demonstrates GT-GAN's superior robustness under severe data loss conditions.}

\section{Conclusion}
This paper addresses the important gap of incomplete or missing data in the growing network measurement datasets for LEO satellite networks. We have proposed a GAN-based framework and comprehensively evaluated the quality of synthetic data generated by three of the latest models in the proposed data missing scenarios. Our findings indicate that when the missing portion of the input data is low, all models perform well in generating synthetic data. However, as the data missing intensifies, a performance gap emerges. Specifically, GT-GAN shows a superior advantage in performance metrics. Even under extreme data missing conditions (e.g., 40\% of the input data is missing), GT-GAN exhibits the highest robustness, consistently capturing the underlying input data distribution and being the least affected in terms of generalization. Our results show the effectiveness of our proposed GAN-based framework and GT-GAN as the most robust and reliable solution for addressing the missing data challenge in LEO satellite network Internet observations. 

{While GT-GAN demonstrates strong overall robustness, it still experiences mode collapse under severe missing rates. To address this problem, we plan to incorporate Wasserstein loss with gradient penalty and apply penalty diversity loss, which may stabilize the training process, effectively prevent the generator from being overpowered by the discriminator, and preserve the diversity of the synthesized data.}


\bibliographystyle{IEEEtran} 
\bibliography{citation}

\end{document}